%% file: mplug.tex
\pdfoutput=1

\documentclass[11pt]{article}

\usepackage{emnlp2021}

\usepackage{times}
\usepackage{latexsym}
\usepackage{microtype}
\usepackage{graphicx}
\usepackage{multirow}
\usepackage[normalem]{ulem}
\useunder{\uline}{\ul}{}
\usepackage{booktabs}
\setcitestyle{square,comma,numbers,sort&compress}

\usepackage{subfig}
\usepackage[linesnumbered,ruled,vlined]{algorithm2e}

\usepackage{listings}
\usepackage{color}

\usepackage{xcolor}

\definecolor{codegreen}{rgb}{0,0.6,0}
\definecolor{codegray}{rgb}{0.5,0.5,0.5}
\definecolor{codepurple}{rgb}{0.58,0,0.82}
\definecolor{backcolour}{rgb}{1.0,1.0,1.0}

\lstdefinestyle{mystyle}{
    backgroundcolor=\color{backcolour},   
    commentstyle=\color{codegreen},
    keywordstyle=\color{magenta},
    numberstyle=\tiny\color{codegray},
    stringstyle=\color{codepurple},
    basicstyle=\ttfamily\small,
    breakatwhitespace=false,         
    breaklines=true,                 
    captionpos=b,                    
    keepspaces=true,                 
    showspaces=false,                
    showstringspaces=false,
    showtabs=false,                  
    tabsize=2
}

\newcommand{\modelname}{mPLUG }

\title{mPLUG: Effective and Efficient Vision-Language Learning by Cross-modal Skip-connections}

\author{Chenliang Li\thanks{\hspace{2mm}Equal contribution}\hspace{1.5mm}, Haiyang Xu$^*$, Junfeng Tian, Wei Wang, Ming Yan\thanks{$^{\dagger}$ Corresponding authors}\hspace{1.5mm}, Bin Bi$^{\dagger}$, Jiabo Ye, Hehong Chen,\\\textbf{Guohai Xu, Zheng Cao, Ji Zhang, Songfang Huang, Fei Huang, Jingren Zhou, Luo Si} \\
  DAMO Academy, Alibaba Group \\
  {\small \texttt{\{lcl193798, shuofeng.xhy, tjf141457, hebian.ww, ym119608, b.bi, yejiabo.yjb, hehong.chh,}}\\
  {\small \texttt{guohai.xgh, zhengzhi.cz, zj122146, songfang.hsf, f.huang, jingren.zhou, luo.si\}@alibaba-inc.com}} \\}

\begin{document}
\maketitle
\begin{abstract}
Large-scale pretrained foundation models have been an emerging paradigm for building artificial intelligence (AI) systems,  which can be quickly adapted to a wide range of downstream tasks. This paper presents mPLUG, a new vision-language foundation model for both cross-modal understanding and generation. Most existing pre-trained models suffer from the problems of low computational efficiency and information asymmetry brought by the long visual sequence in cross-modal alignment. To address these problems, \modelname introduces an effective and efficient vision-language architecture with novel cross-modal skip-connections, which creates inter-layer shortcuts that skip a certain number of layers for time-consuming full self-attention on the vision side.

\modelname is pre-trained end-to-end on large-scale image-text pairs with both discriminative and generative objectives. It achieves state-of-the-art results on a wide range of vision-language downstream tasks, such as image captioning, image-text retrieval, visual grounding and visual question answering. \modelname also demonstrates strong zero-shot transferability when directly transferred to multiple video-language tasks. 

\end{abstract}


\section{Introduction} 
Large-scale pre-training of vision-language models have recently received tremendous success on a wide range of cross-modal tasks~\cite{tan2019lxmert,chen2020uniter,huang2020pixel,li2020oscar,yu2021ernie,li2021align,wang2021simvlm}. Such vision-language models learn cross-modal representations from a quantity of image-text pairs by aligning the visual and linguistic modalities. A great challenge of learning vision-language models is to find a good alignment between the two modalities to close the semantic gap in-between.

\begin{figure}[t]
     \centering
     \includegraphics[width=0.5\textwidth]{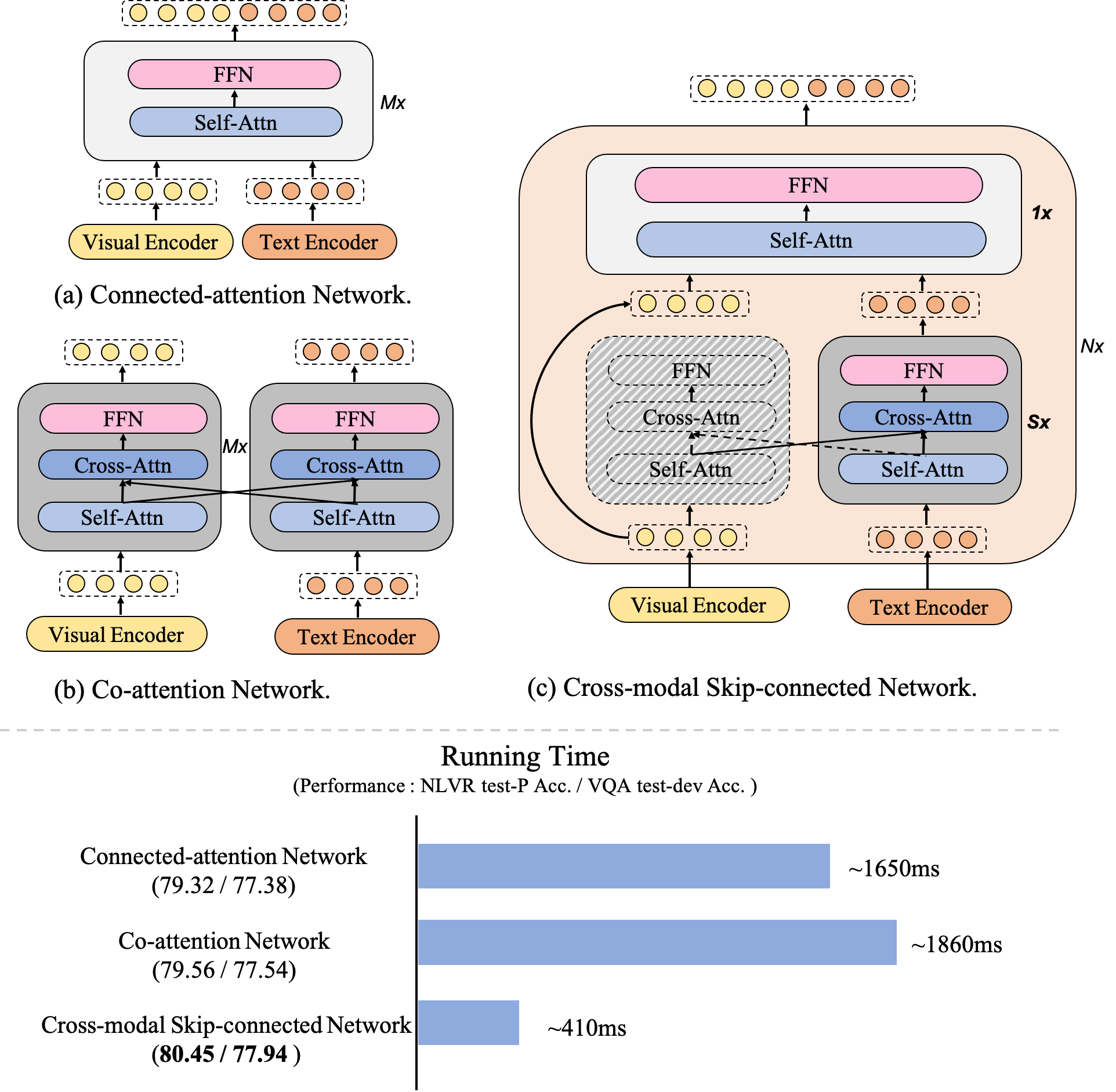}
     \caption{Illustration of two conventional cross-modal fusion networks and our proposed cross-modal skip-connected network. We compare the running time and performance of different fusion networks, where the total fusion layers, image encoder and text encoder are all kept the same. The running time is the total forward time of 100 samples in different fusion networks.}
     \label{fig:compare}
\end{figure}
To discover a cross-modal alignment, prior studies~\cite{li2020oscar,anderson2018bottom,zhang2021vinvl} employ a pre-trained object detector to extract salient regions from images, which are then aligned with language counterparts. Such an architecture, however, is generally limited by the power of the object detector, the pre-defined visual semantics it can represent, and the quantity of annotations available. Besides, it is also computationally expensive to extract region-based visual features from high-resolution (e.g. 600$\times$1000) images. More recent work~\cite{huang2020pixel,wang2021simvlm,li2021align,kim2021vilt,dou2021empirical}, which scales and performs better on many vision-language tasks, drops the requirement of pre-trained object detection and enables a direct alignment between the image and text representations in an end-to-end manner. These models extract finer-grained visual representation with a long sequence of image patches or grids for good vision understanding~\cite{dou2021empirical}. However, there exist two significant problems in modeling long visual sequences: 1) \textit{efficiency}: full self-attention on long visual sequences requires much more computation than that on textual sequences, and 2) \textit{information asymmetry}: the caption text in widely-used image-text pre-training data is usually short and highly abstract while more detailed and diverse information can be extracted from the image. This asymmetry presents challenges for effective multi-modal fusion between the modalities.

One straightforward way of multi-modal fusion is the connected-attention network as shown in Figure~\ref{fig:compare} (a). It adopts a single Transformer~\cite{vaswani2017attention} network for early fusion of vision and language by simply taking the concatenation of visual and linguistic features as input~\cite{li2019visualbert}. This paradigm allows self-attention to discover alignments between the modalities from the bottom level, and requires full self-attention on the concatenation of cross-modal sequences, which is rather time-consuming. Besides, this type of methods process information from both modalities equally, which may suffer from the information asymmetry especially when there is a big difference in information density or sequence lengths between the modalities.


Another line of work keeps separate Transformer networks for both textual and visual features, and uses techniques such as cross-attention to enable cross-modal interaction~\cite{dou2021empirical}, as shown in Figure~\ref{fig:compare} (b). This architecture design conducts multi-modal fusion on both modalities independently, which can help alleviate the information asymmetry problem. However, it still suffers from computation inefficiency for full self-attention on long visual sequences, and it is not that parameter-efficient with two separate Transformer networks.


In this work, we propose mPLUG, a unified Multi-modal Pre-training framework for both vision-Language Understanding and Generation. \modelname performs effective and efficient vision-language learning with novel cross-modal skip-connections to address the fundamental information asymmetry problem. Instead of fusing visual and linguistic representations at the same levels, the cross-modal skip-connections enables the fusion to occur at disparate levels in the abstraction hierarchy across the modalities. It creates inter-layer shortcuts that skip a certain number of layers for visual representations to reflect the semantic richness of language compared to vision. As shown in Figure~\ref{fig:compare} (c), in each block of our cross-modal skip-connected network, \modelname first adopts an asymmetric co-attention architecture at the first few layers for efficiency, by removing the co-attention on vision side. It is then followed by one layer of connected-attention, by concatenating the original visual representation and the co-attention output on the language side as input. In addition to the modeling efficacy due to the asymmetry, the cross-modal skip-connections ease the model training by alleviating vanishing gradients with the inserted shortcuts. Figure~\ref{fig:compare} shows that the new cross-modal skip-connected network achieves superior performance with at least four times speeding-up than other cross-modal fusion networks.

Our key contributions can be summarized as follows:
\begin{itemize}
\item
We propose a unified vision-language pretrained model \modelname of cross-modal understanding and generation for both effectiveness and efficiency in cross-modal learning.
\item 
We introduce a new asymmetric vision-language architecture with novel cross-modal skip-connections, to address two fundamental problems of information asymmetry and computation inefficiency in multi-modal fusion.
\item 
\modelname achieves state-of-the-art performance on a wide range of vision-language tasks, including image captioning, image-text retrieval, visual grounding and visual question answering. \modelname also demonstrates strong zero-shot transferability when directly transferred to a wide range of vision-language and video-language tasks.

\end{itemize}


\section{Related Work} 
\subsection{Vision-Language Pre-training}
Vision-Language pre-training (VLP) has recently received tremendous success and achieved state-of-the-art results across a variety of vision-language  tasks~\cite{antol2015vqa,chen2015microsoft,yu2016modeling}. In terms of how information from different modalities are aggregated, typical approaches to VLP~\cite{tan2019lxmert,chen2020uniter,huang2020pixel,yu2021ernie,li2021align,radford2021learning,jia2021scaling} can be roughly divided into two categories: \textit{dual encoder} and \textit{fusion encoder}. Dual encoder approach utilizes two single-modal encoders to encode images and text separately, and then uses simple functions such as dot product to model the instance-level cross-modal interaction between image and text. The advantage of dual encoder models like CLIP~\cite{radford2021learning} and ALIGN~\cite{jia2021scaling} is that images and text can be pre-computed and cached, which is quite computation-efficient and more appropriate for retrieval tasks. However, they tend to fail in handling more complicated VL understanding tasks that require complex reasoning, such as visual question answering~\cite{antol2015vqa}. In contrast, fusion encoder approach uses deep fusion functions such as multi-layer self-attention and cross-attention networks to model the fine-grained cross-modal interaction between image and text sequences. Representative methods of this category include the single-stream architecture such as UNITER~\cite{chen2020uniter} and OSCAR~\cite{li2020oscar}, and two-stream architecture such as LXMERT~\cite{tan2019lxmert}, ALBEF~\cite{li2021align} and ERNIE-ViL~\cite{yu2021ernie}. This kind of methods can better capture the underlying association between image and text for vision-language understanding tasks, while it needs to jointly encode all possible image-text pairs, which leads to a relatively slow inference speed. 

To improve the inference speed, some recent work such as Pixel-BERT~\cite{huang2020pixel}, E2E-VLP ~\cite{xu2021e2e} and ViLT~\cite{kim2021vilt} removes the complicated object detector in feature extraction, and conducts end-to-end VL learning with CNN-based grid features and linearly projected patched embeddings, respectively. To combine the benefits of both categories of architectures, VLMo~\cite{wang2021vlmo} further unifies the dual encoder and fusion encoder modules with shared mixture-of-modality-experts Transformer. In this work, mPLUG introduces a new cross-modal fusion mechanism with cross-modal skip-connections, to enables the fusion to occur at disparate levels in the abstraction hierarchy across the modalities. It achieves superior performances in effectiveness and efficiency across a wide range of VL tasks.

\subsection{Skip-connection}
Skip-connection is a popular technique to bypass the gradient exploding or vanishing problem for model optimization in deep neural networks, which is widely-used in CV and NLP architectures such as ResNet~\cite{he2016deep} and Transformer~\cite{vaswani2017attention}. A variety of skip connection methods have been proposed in recent years~\cite{srivastava2015highway,he2016deep,vaswani2017attention,huang2017densely,szegedy2017inception,liu2021rethinking}. ResNet~\cite{he2016deep} introduces summed shortcut connections between different layers using simple identity mapping, while highway network~\cite{srivastava2015highway} designs a transform gating function to control the balance of the input and the transformed input. DenseNet~\cite{huang2017densely} designs new architectures with concatenated skip-connections, allowing the subsequent layers to re-use all the middle representations of previous layers. Layer Normalization and recursive skip connection are further used in combination with plain skip connection for further stablizing model optimization and better incorporating the transformed input~\cite{vaswani2017attention,liu2021rethinking}. In this work, mPLUG proposes a new cross-modal skip connection method to address cross-modal fusion problem, and combines the concatenated skip-connection and summed skip-connection for choosing whether to attend to all the concatenated representations of different modalities or just focus on the cross-modal interaction part at each layer. 


\begin{figure*}
     \centering
     \subfloat{\includegraphics[width=0.5\textwidth]{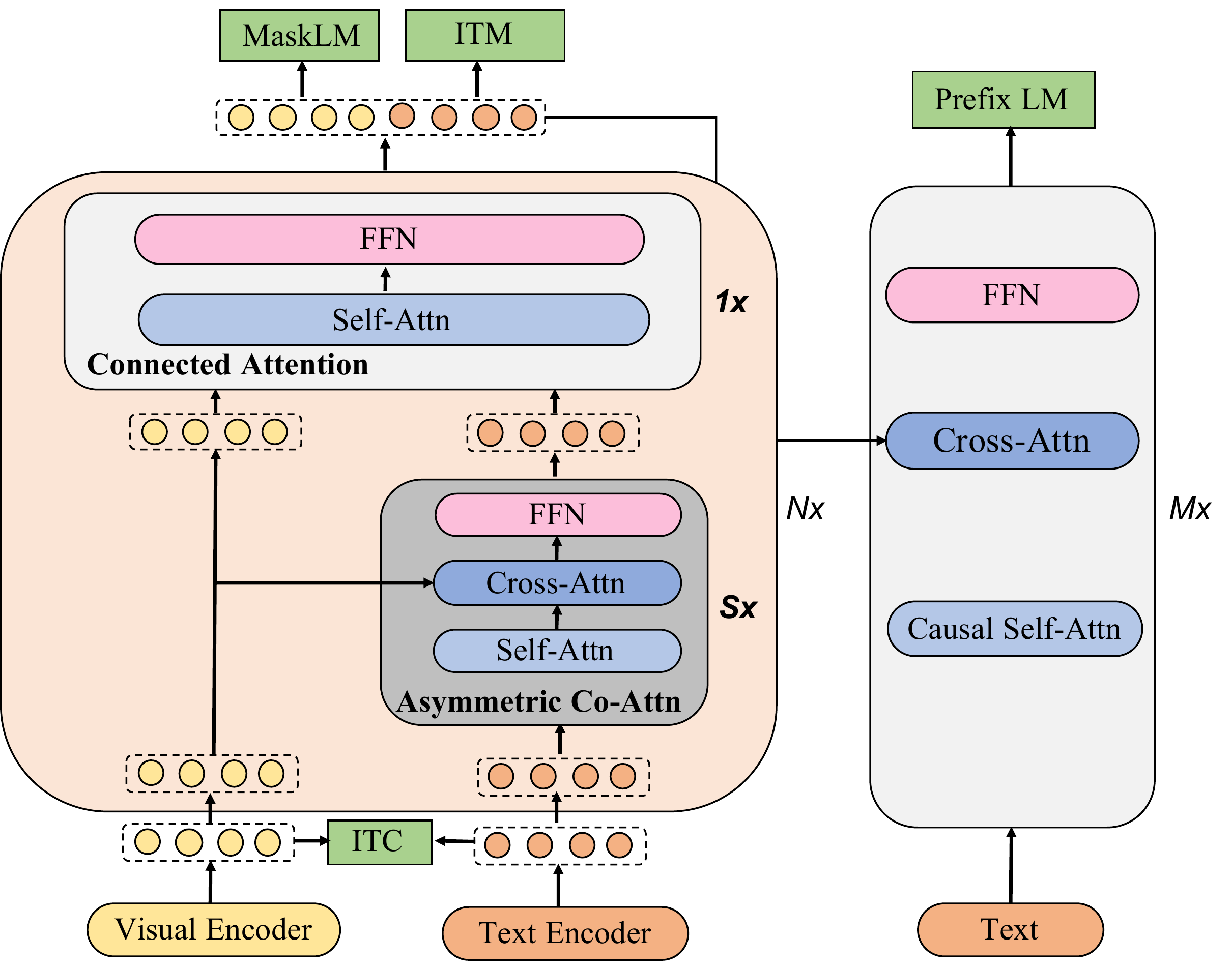}} 
     \subfloat{\includegraphics[width=0.44\textwidth]{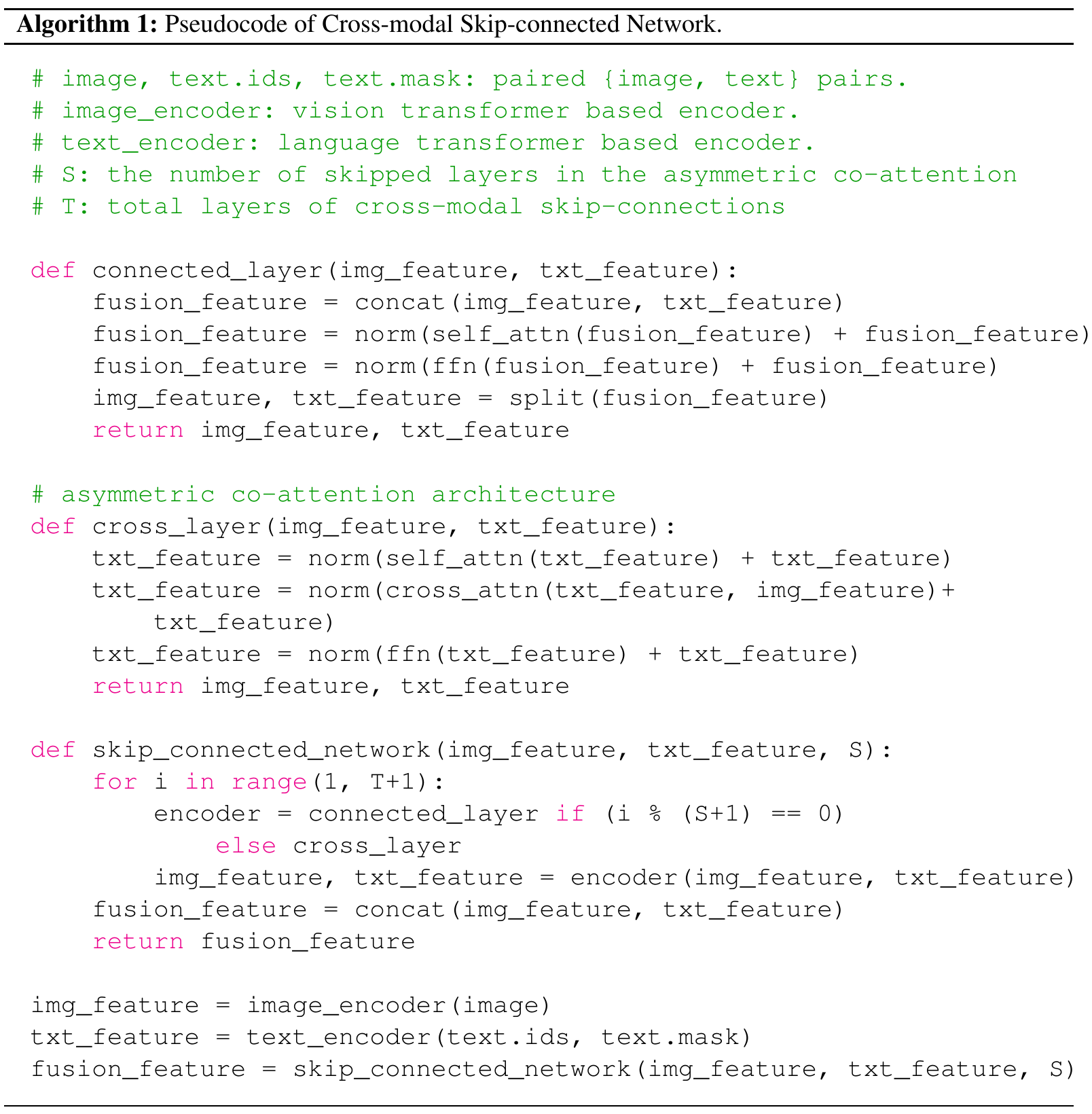}} 
     \caption{The model architecture and objectives of mPLUG, which consists of two unimodal encoders for images and text separately, a cross-modal skip-connected network and a decoder for text generation. An image-text contrastive loss is first applied to align the unimodal representations from the visual encoder and text encoder. Then, we use a novel cross-modal skip-connected network to fuse the visual and linguistic representations effectively and efficiently. We adopt connected cross-modal fusion to every $S$ \textit{asymmetric co-attention} layers, where $S$ is a fixed stride value. Based on the connected representation of the image and prefix sub-sequence, the decoder is trained with a prefix language modeling (Prefix LM) loss by generating the remaining caption.}
     \label{fig:framework}
\end{figure*}

\section{\modelname}
In this section, we will first introduce our new model architecture with the key module of the cross-modal skip-connected network, and then give the details of the pre-training objectives and scalable training infrastructure. 

\subsection{Model Architecture}

As shown in Figure \ref{fig:framework}, \modelname consists of two unimodal encoders for image and text independently, a cross-modal skip-connected network and a decoder for text generation. To better model the inherent modality bias information, we first use two unimodal encoders to encode image and text separately. Following \cite{dou2021empirical,shen2021much}, we use a visual transformer~\cite{dosovitskiy2020image} directly on the image patches as the visual encoder, which is more computation-friendly than using pre-trained object detectors for visual feature extraction~\cite{anderson2018bottom,zhang2021vinvl}. The visual encoder divides an input image into patches and encodes them as a sequence of embeddings $\{v_{cls}, v_1,v_2,...,v_M\}$ with an additional $[CLS]$ token. The input text is fed to the text encoder and represented as a sequence of embeddings $\{l_{cls}, l_1, l_2,...,l_N\}$, where $l_{cls}$ is the embedding of the $[CLS]$ token and used to summarize the input text. Then, the visual and linguistic representations are fed into a cross-modal skip-connected network, which consists of multiple skip-connected fusion blocks. In each skip-connected fusion block, we adopt connected cross-modal fusion to each of $S$ \textit{asymmetric co-attention}  layers where $S$ is a fixed stride value. The aim of this network is to take advantage of the effectiveness of the connected cross-modal fusion and the efficiency of the asymmetric co-attention for enhanced cross-modal fusion in a recursive manner. Finally, the output cross-modal representations are fed into a transformer decoder for sequence to sequence learning, which equips \modelname with both understanding and generation capabilities.


\subsection{Cross-modal Skip-connected Network}

The cross-modal skip-connected network consists of $N$ skip-connected fusion blocks. In each skip-connected fusion block, we adopt \textit{connected-attention} layer to each of $S$ \textit{asymmetric co-attention} layers where $S$ is a fixed stride value. We first pass the text feature and image feature from unimodal encoders through the $S$ asymmetric co-attention layers, and then connect the output text feature and image feature to one connected-attention layer. We repeat the skip-connected fusion block $N$ times for the final connected image and text representation.

Specifically, the asymmetric co-attention is composed of the self-attention (SA) layer, cross-attention (CA) layer and the feed-forward network (FFN). The input text feature $l^{n-1}$ is first fed to the self-attention layer, and then the visual feature $v^{n-1}$ is injected into the text feature $l^{n}_{SA}$ by the cross-attention layer which gives $l^{n}_{CA}$. The output of self-attention $l^{n}_{SA}$ and cross-attention $l^{n}_{SA}$ are added up and fed to the FFN layer for the visual-aware text representation $l^{n}$:

\begin{equation}
l^{n}_{SA}=LN(SA(l^{n-1})+l^{n-1})
\end{equation}
\begin{equation}
l^{n}_{CA}=LN(CA(l^{n}_{SA},v^{n-1})+l^{n}_{SA})
\end{equation}
\begin{equation}
l^n=LN(FFN(l^{n}_{CA})+l^{n}_{CA})
\end{equation}
where LN is short for layer normalization.

The connected-attention layer is composed of the self-attention (SA) layer and the feed-forward network (FFN). We connect the image feature $v^{n-1}$ and input text feature $l^{n-1}$, where $l^{n-1}$ is the output of $S$ asymmetric co-attention layers. The connected image and text feature $[v^{n-1};l^{n-1}]$ are fed to the self-attention layer and FFN layer:
\begin{equation}
[v^n_{SA};l^n_{SA}]=LN(SA([v^{n-1};l^{n-1}])+[v^{n-1};l^{n-1}])
\end{equation}
\begin{equation}
[v^n;l^n]=LN(FFN([v^n_{SA};l^n_{SA}])+[v^n_{SA};l^n_{SA}])
\end{equation}

Then $[v^n;l^n]$ is fed into the next cross-modal skip-connected network repeatedly to get the final connected image and text representation. Finally, the connected output is fed into a Transformer decoder for sequence to sequence learning.

\subsection{Pre-training Tasks}
We perform four pre-training tasks including three understanding tasks (Image-Text Contrastive Learning, Image-Text Matching, Masked Language Modeling) and one generation task (Prefix Language Modeling). 
These pre-training tasks are optimized jointly.

\textbf{Image-Text Contrastive (ITC)}: Following ~\cite{li2021align}, we employ the task to align the image features and the text features from the unimodal encoders. Specifically, we calculate the softmax-normalized image-to-text and text-to-image similarity, and take two dynamic memory queues (text, image) to increase the number of negative examples as MoCo ~\cite{he2020momentum}.

\textbf{Image-Text Matching (ITM)}: This task aims to predict whether an image and a sentence match with each other on the cross-modal representation. We also select hard negative image-text pairs based on the contrastive text-image similarity as ~\cite{li2021align}.

\textbf{Masked Language Modeling (MLM)}: The task setup is basically the same as in BERT~\cite{devlin2018bert}, where we randomly mask $15\%$ of tokens in text and the model is asked to predict these masked words with the cross-modal representations.

\textbf{Prefix Language Modeling (PrefixLM)}: This task aims to generate the caption given an image and predict the text segment subsequent to the cross-modal context as ~\cite{bi2020palm}. It optimizes a cross entropy loss
by maximizing the likelihood of text in an autoregressive manner.

\section{Distributed Learning on a Large Scale} 
Training a big model like mPLUG on large-scale datasets faces many efficiency challenges. We increase the throughput from the perspective of reducing memory usage and computation time, thereby accelerating the training of the model.

The memory usage during model training is mainly composed of two aspects: the static memory usage composed of parameters/optimizer states/gradients, etc., and the runtime memory usage caused by intermediate variables like activation values.
For static memory overhead, we use the ZeRO~\cite{rajbhandari2020zero} technique to partition parameters/optimizer states/gradients into the entire data-parallel group, so that the static memory overhead of a single GPU can be approximately reduced to $1/N$, where $N$ denotes the number of GPU cards.
We use gradient checkpointing~\cite{chen2016training} for the runtime memory cost, which greatly reduces the runtime memory usage at the expense of increasing forward time by recomputing part of the activation values during backward pass without keeping them in memory.

To reduce the computation time, we use BF16 precision training. BF16 is a new data type supported by NVIDIA's new Ampere architecture GPU like A100. Compared with the previously widely used mixed-precision training of FP16 and FP32, BF16 has the same representation range as FP32, thereby reducing the risk of numerical overflow and ensuring model convergence stability, and at the same time has the same fast computing speed as FP16.


\section{Experiments} 
\subsection{Data \& Setup}

\begin{table*}
\setlength\tabcolsep{4pt}
\centering
\begin{tabular}{l|c|cccccccc|cc}
\toprule
\multicolumn{1}{c|}{\multirow{3}{*}{Models}}      &
\multicolumn{1}{c|}{\multirow{3}{*}{Data}} &
\multicolumn{8}{c|}{COCO Caption} & \multicolumn{2}{c}{\multirow{1}{*}{NoCaps}}  \\
\multicolumn{1}{c|}{\multirow{2}{*}{}}      &
\multicolumn{1}{c|}{} &
\multicolumn{4}{c}{Cross-entropy Optimization} & \multicolumn{4}{c|}{CIDEr Optimization} & \multicolumn{2}{c}{}  \\
      &  & B@4 & M & C & S & B@4 & M & C & S & C & S     \\
      
\midrule      
Encoder-Decoder & CC12M & - & - & 110.9 & - &  - & - & - & - & 90.2 & 12.1 \\
E2E-VLP \cite{xu2021e2e} &4M& 36.2 &-&117.3&-&  - & - & - & - & - & - \\
VinVL \cite{zhang2021vinvl} & 5.65M & 38.5 & 30.4 & 130.8 & 23.4 & 41.0 & 31.1 & 140.9 & 25.2 & 97.3 & 13.8 \\
OSCAR \cite{li2020oscar} & 6.5M & - & - & - & - & 41.7 & 30.6 & 140.0 & 24.5 & 83.4 & 11.4 \\
SimVLM$_{large}$ \cite{wang2021simvlm} & 1.8B & 40.3 & \textbf{33.4} & \textbf{142.6} & \textbf{24.7} & - & - & - & - & - & - \\
LEMON$_{large}$ \cite{LEMON} & 200M & 40.6 & 30.4 & 135.7 & 23.5 & 42.3 & 31.2 & 144.3 & 25.3 & 113.4 & \textbf{15.0} \\
BLIP \cite{li2022blip} & 129M & 40.4 & - & 136.7 & - & - & - & - & - & 113.2 & 14.8  \\
OFA \cite{wang2022OFA} & 18M & - & - & - & - & 43.5 & 31.9 & 149.6 & \textbf{26.1} & - & - \\
\midrule
mPLUG & 14M  & \textbf{43.1} & 31.4 & 141.0 & 24.2 & \textbf{46.5} & \textbf{32.0} & \textbf{155.1} & 26.0 & \textbf{114.8} & 14.8 \\
\midrule

\end{tabular}
\caption{Evaluation Results on COCO Caption "Karpathy" test split and NoCaps validation set. B@4: BLEU@4, M: METEOR, C: CIDEr, S: SPICE.} 
\label{table:caption}
\end{table*}

Following the previous work ~\cite{li2021align}, we use the same pre-training dataset with 14M images with texts, which includes two in-domain datasets (MS COCO ~\cite{lin2014microsoft} and Visual Genome ~\cite{krishna2017visual}), and
three web out-domain datasets (Conceptual Captions ~\cite{sharma2018conceptual}, Conceptual 12M ~\cite{changpinyo2021conceptual}, SBU Captions ~\cite{ordonez2011im2text}. 

We pretrain the model for 30 epochs with the total batch size of 1024 on 16 NVIDIA A100 GPUs. We use a 6-layer Transformer for both the text encoder and the cross-modal skip-connected network, and a 12-layer Transformer for the decoder. The text encoder is initialized using the first 6 layers of the  BERT$_{base}$~\cite{devlin2018bert} model and the skip-connected network is initialized using the last 6 layers of the~BERT$_{base}$. We initialize the visual encoder by CLIP-ViT ~\cite{radford2021learning} pretrained on 400M noisy image-text pairs. The visual transformer with ViT-B/16 is used as our base architecture, the one with ViT-L/14 as the large architecture. We use the AdamW ~\cite{loshchilov2017decoupled} optimizer with a weight decay of 0.02. The learning rate is warmed-up to 1e-5 (ViT-B/16) and 1e-4 (BERT$_{base}$) for mPLUG{\tiny{ViT-B}} , and 5e-6 (ViT-L/14) and 5e-5 (BERT$_{base}$) for mPLUG{\tiny{ViT-L}} in the first 1000 iterations, and decayed to 1e-6 following a cosine schedule. During pre-training, we take random image crops of resolution 256 $\times$ 256 (ViT-B/16)/224 $\times$ 224 (ViT-L/14) as input, and also apply RandAugment ~\cite{cubuk2020randaugment} to improve the generalization of vision encoders. For VQA and image captioning tasks, we do an additional  continue pre-training on 4M image-text pairs.
We increase the image resolution during finetuning. For image-text contrastive learning, the queue size is set as 65,536 and the momentum coefficient is set as 0.995. 
\subsection{Evaluation on Vision-Language Tasks}
We compare our pre-trained model against other VLP models on the six downstream V+L tasks. We introduce each task and our fine-tuning strategy below. 
Details of the datasets and fine-tuning hyperparameters are in Appendix.

\begin{table}[t]
\setlength\tabcolsep{3pt}
\centering
\begin{tabular}{lccc}
\toprule
Models & Data & Test-dev & Test-std \\
\midrule
\multicolumn{4}{l}{\emph{Pretrained on COCO, VG, SBU and CC datasets}} \\
VLBERT \cite{lu2019vilbert} & 4M   & 71.16    & -   \\
E2E-VLP \cite{xu2021e2e} & 4M & 73.25  & 73.67 \\
VL-T5 \cite{vlt5} & 4M & - & 71.30 \\
UNITER\cite{chen2020uniter} & 4M     & 72.70    & 72.91   \\
OSCAR\cite{li2020oscar} & 4M  & 73.16   & 73.44   \\
CLIP-ViL\cite{shen2021much} & 4M & 76.48 & 76.94 \\
METER\cite{dou2021empirical} & 4M & 77.68 & 77.64 \\
ALBEF\cite{li2021align} & 4M & 74.54 & 74.70   \\
mPLUG\tiny{ViT-B} & 4M  & \textbf{77.94} & \textbf{77.96} \\

\midrule
\multicolumn{4}{l}{\emph{Models Pretrained on More Data}} \\
ALBEF~\cite{li2021align} & 14M & 75.84 & 76.04 \\
BLIP~\cite{li2022blip} & 129M & 78.25 & 78.32 \\
SimVLM~\cite{wang2021simvlm} & 1.8B  &  80.03 & 80.34 \\
Florence~\cite{yuan2021florence} & 0.9B & 80.16&80.36\\
OFA~\cite{wang2022OFA} & 18M & 79.87 & 80.02 \\
VLMo~\cite{wang2021vlmo} & - & 79.94 & 79.98 \\
mPLUG\tiny{ViT-B} & 14M   & 79.79   & 79.81 \\
mPLUG\tiny{ViT-L} & 14M   & \textbf{81.27}  & \textbf{81.26} \\
\bottomrule
\end{tabular}
\caption{Evaluation Results on VQA test set.}
\label{table:vqa}
\end{table}

\subsubsection{Visual Question Answering} \label{section:vqa_analysis} 
The VQA task ~\cite{antol2015vqa} requires the model to answer natural language questions given an image. Most methods~\cite{tan2019lxmert,wang2021vlmo,li2020oscar,wang2021simvlm} deal with visual question answering tasks as multi-label classification on predefined answer sets. This strategy achieves strong performance, but it is not suitable for real-world open scenarios. We treat VQA as an answer generation task and directly use unconstrained open-vocab generation during inference, which is different from constrained close-vocab generation models \cite{li2021align,wang2022OFA}. Following \cite{li2020oscar,wang2022OFA}, we concatenate the question with the object labels and OCR tokens extracted from image. As shown in Table \ref{table:vqa}, \modelname achieves 81.27 on Test-std split and outperforms the SOTA models including SimVLM and Florence, which use 100$X$ and 60$X$ more pre-training image-text pairs, respectively. Based on the same 4M pre-training data, \modelname outperforms CLIP-ViL and METER, which also use CLIP~\cite{radford2021learning} as the visual encoder. Besides, under the same settings, \modelname always significantly outperforms ALBEF and BLIP which only rely on co-attention from images to text for cross-modal fusion. The gain can derive from the network design of cross-modal skip-connections specifically for information asymmetry of the two modalities. Neither ALBEF nor BLIP addresses this problem well, with bias towards the language modality.

\begin{table*}
\setlength\tabcolsep{4pt}
\centering
\small
\begin{tabular}{l|c|cccccc|cccccc}
\toprule
\multicolumn{1}{c|}{\multirow{2}{*}{Models}}      &
\multicolumn{1}{c|}{\# Pretrain} &
\multicolumn{6}{c|}{MSCOCO (5K test set)} & \multicolumn{6}{c}{Flickr30K (1K test set)} \\
      &  data & \multicolumn{3}{c}{TR} & \multicolumn{3}{c|}{IR} & \multicolumn{3}{c}{TR} & \multicolumn{3}{c}{IR}          \\
\midrule
&&R@1&R@5&R@10&R@1&R@5&R@10&R@1&R@5&R@10&R@1&R@5&R@10 \\
E2E-VLP~\cite{xu2021e2e}& 4M     &-& -&-&-&-&- & 86.2 &97.5 &98.92&73.6 & 92.4 &96.0 \\
UNITER~\cite{chen2020uniter} & 4M     & 65.7&88.6&93.8&52.9&79.9&88.0&87.3& 98.0&99.2&75.6&94.1&96.8  \\
OSCAR~\cite{li2020oscar} & 4M  & 70.0&91.1&95.5&54.0&80.8&88.5&-& -&-&-&-&-   \\
UNIMO ~\cite{li2020unimo} & 4M     &-& -&-&-&-&- & 89.4 & 98.9& 99.8 &78.0 &94.2& 97.1\\
VLMo ~\cite{wang2021vlmo} & 4M & 78.2& 94.4& 97.4& 60.6& 84.4& 91.0& 95.3& 99.9& 100.0& 84.5& 97.3& 98.6 \\
ALIGN~\cite{jia2021scaling} & 1.8B  & 77.0&93.5&96.9&59.9&83.3&89.8&95.3& 99.8&100.0&84.9&97.4&98.6   \\
ALBEF ~\cite{li2021align} & 14M & 77.6&94.3&97.2&60.7&84.3&90.5&95.9& 99.8&100.0&85.6&97.5&98.9                 \\
Florence ~\cite{yuan2021florence} & 0.9B & 81.8&95.2&-&63.2&85.7&-&97.2& 99.9&-&87.9&98.1&-                 \\
BLIP ~\cite{li2022blip}& 14M & 80.6 &95.2&97.6&63.1&85.3&91.1&96.6& 99.8&100.0&87.2&97.5&98.8                 \\
BLIP ~\cite{li2022blip}& 129M & 82.4 &95.4&97.9&65.1&86.3&91.8&97.4& 99.8&99.9&87.6&97.7&99.0                 \\
\midrule
mPLUG & 14M  & \textbf{82.8} &\textbf{96.1}&\textbf{98.3}&\textbf{65.8}&\textbf{87.3}&\textbf{92.6}&\textbf{97.6}& \textbf{100.0}&\textbf{100.0}&\textbf{88.4}&\textbf{97.9}&\textbf{99.1}  \\
\bottomrule
\end{tabular}          \\
\caption{Image-text retrieval results on Flickr30K and COCO datasets.}
\label{table:retrieval}
\end{table*}

\subsubsection{Image Captioning} 
The image captioning task requires a model to generate an appropriate and fluent caption for a given image. We evaluate image captioning on two datasets COCO Caption~\cite{coco_caption} and NoCaps~\cite{nocaps}. \modelname finetuned with training data of COCO Caption is tested on both of the datasets. We train \modelname on the MS COCO Caption and test on the same Karpathy split~\cite{li2020oscar,wang2021simvlm} and NoCaps validation set. Following~\cite{li2020oscar,wang2022OFA}, we first fine-tune \modelname with cross-entropy loss and then with CIDEr optimization~\cite{scst} for extra 5 epochs. As shown in Table~\ref{table:caption}, \modelname with only 14M pre-training images can outperform the SOTA models including LEMON and SimVLM on both COCO Caption and Nocaps datasets, which uses more than 10$X$ and 100$X$ pre-training data, respectively. For the COCO Caption, \modelname performs the best on CIDEr evaluation and surpasses the SOTA model by a large margin of 5.5 on Karpathy test set. We use the best checkpoint on COCO Caption and predict on the Nocaps validation set directly.


\begin{table*}[t]
\centering
\begin{tabular}{@{}lcccccccc@{}}
\toprule
\multicolumn{1}{c}{\multirow{2}{*}{Model}} & \multicolumn{3}{c}{RefCOCO} & \multicolumn{3}{c}{RefCOCO+} & \multicolumn{2}{c}{RefCOCOg} \\
\multicolumn{1}{c}{}                       & val     & testA   & testB   & val      & testA   & testB   & val-u        & test-u        \\ \midrule
VLBERT~\cite{lu2019vilbert}\                                    & -       & -       & -       & 72.59    & 78/57   & 62.30   & -            & -             \\
UNITER~\cite{chen2020uniter}                                     & 81.41   & 87.04   & 74.17   & 75.90    & 81.45   & 66.70   & 74.86        & 75.77         \\
VILLA~\cite{DBLP:conf/nips/Gan0LZ0020}                                    & 82.39   & 87.48   & 74.84   & 76.17    & 81.54   & 66.84   & 76.18        & 76.71         \\
MDETR~\cite{DBLP:conf/iccv/KamathSLSMC21}                                      & 86.75   & 89.58   & 81.41   & 79.52    & 84.09   & 70.62   & 81.64        & 80.89         \\
UNICORN~\cite{DBLP:journals/corr/abs-2111-12085}                                    & 88.29   & 90.42   & 83.06   & 80.30    & 85.05   & 71.88   & 83.44        & 83.93         \\
OFA~\cite{wang2022OFA}                                        & 90.05   & 92.93   & 85.26   & 84.49    & 90.10   & 77.77   & 84.54        & 85.20         \\ \midrule
mPLUG                                 & \textbf{92.40}   & \textbf{94.51}   & \textbf{88.42}   &  \textbf{86.02}   &  \textbf{90.17}   &  \textbf{78.17}   &    \textbf{85.88}   & \textbf{86.42}              \\ \bottomrule
\end{tabular}

\caption{Visual grounding results (Acc@0.5) on ReferCOCO, ReferCOCO+, and ReferCOCOg.}
\label{tab:visual_grounding}
\end{table*}

\subsubsection{Image-Text Retrieval} 
We conduct experiments for both image-to-text retrieval (TR) and text-to-image retrieval (IR) on COCO ~\cite{lin2014microsoft} and Flickr30K ~\cite{plummer2015flickr30k} datasets. Following ~\cite{li2021align, li2022blip}, we jointly optimize the ITC loss and the ITM loss during fine-tuning. During inference, we first select top-k candidates by computing the dot-product similarity between the image and text encoder features, and then rerank
the selected candidates based on their ITM scores. We set $k = 256$ for COCO and $k = 128$ for Flickr30K. As shown in Table \ref{table:retrieval}, \modelname outperforms all existing methods on both datasets. Using
14M images, \modelname achieves better performance than
BLIP with 129M and Florence with 0.9B pre-training data. Using the same 14M pre-training images, \modelname substantially outperforms the previous best model BLIP by +2.7\% in TR recall@1 on COCO and +1.0 \% in TR recall@1 on Flickr30K.

\subsubsection{Visual Grounding} 
Given a query in plain text and an image, visual grounding requires models to localize the referred object in the image. Instead of regressing the bounding boxes directly, we concatenate visual features and attended textual features and feed them into the decoder to predict the coordinates. Table \ref{tab:visual_grounding} shows that \modelname outperforms all the SOTA methods. We observe that in RefCOCO testB the images often contain arbitrary objects and in RecCOCOg test-u the expressions are longer than other datasets. Compared with the previous best model OFA, \modelname achieves 3.16\% absolute improvement on RefCOCO testB and 1.22\% absolute improvement on RefCOCOg test-u. It demonstrates that \modelname learns better multi-modal interaction from cross-modal skip-connections and is better at handling complex images and long queries.

\subsubsection{Visual Reasoning}
We consider two datasets for visual reasoning: NLVR2~\cite{suhr2018corpus} and SNLI-VE~\cite{snlive}. The NLVR2~\cite{suhr2018corpus} task requires the model to predict whether a sentence describes a pair of images. Following \cite{li2022blip}, we use two cross-attention layers to process the two input images, and their outputs are merged and fed to the FFN. An MLP classifier is then applied on the output embedding of the language [CLS] token. The SNLI-VE~\cite{snlive} task requires the model to evaluate how the given image and text are semantically correlated, i.e., entailment, neutral, or contradiction. Following \cite{wang2022OFA}, the image premise, text premise and text hypothesis are fed to the encoder. While we remove the decoder, and only use the encoder modules for three-way classification, which can save nearly half of the total computation cost. We predict the class probabilities using the multimodal encoder's output representation of the language [CLS] token. As shown in Table \ref{table:reasoning}, \modelname can obtain competitive performances to the SOTA models~\footnote{The SOTA models  such as OFA and VLMo both add large-scale text-only and image-only pre-training data for improving the reasoning ability.} in both visual reasoning tasks, and even outperform SimVLM~\cite{wang2021simvlm} and BLIP~\cite{li2022blip}, which use far more pre-training data.

\begin{table}[t]
\setlength\tabcolsep{5pt}
\centering
\begin{tabular}{lcccc}
\toprule
\multicolumn{1}{l}{\multirow{2}{*}{Model}}      &
\multicolumn{2}{c}{NLVR2} & \multicolumn{2}{c}{SNLI-VE} \\
       & dev & test-P &  dev & test   \\
\midrule
LXMERT\cite{tan2019lxmert} & 74.90 & 74.50 & - & - \\
VL-T5\cite{vlt5}  & - & 73.6 & - & - \\
UNITER\cite{chen2020uniter} & 79.12 & 79.98 & 79.39 & 79.38  \\
CLIP-ViL\cite{shen2021much} & - & - & 80.61 & 80.20 \\
METER\cite{dou2021empirical} & 82.33 & 83.05 & 80.86 & 81.19 \\
UNIMO\cite{li2020unimo} & - & - & 81.11 & 80.63  \\ 
ALBEF\cite{li2021align} & 82.55 & 83.14 & 80.80 & 80.91 \\
BLIP\cite{li2022blip}  & 82.67 & 82.30 & - & - \\
SimVLM$_{large}$\cite{wang2021simvlm} & 84.13 & 84.84 & 85.68 & 85.62 \\
VLMo\cite{wang2021vlmo} & \textbf{85.64} & \textbf{86.86} & - & - \\
OFA\cite{wang2022OFA}  & - & - & \textbf{90.30} & \textbf{90.20} \\
\midrule
mPLUG  & 84.58  & 84.95 & 89.45 & 89.29 \\
\bottomrule
\end{tabular}
\caption{Evaluation Results on NLVR2 and SNLI-VE.}
\label{table:reasoning}
\end{table}







\subsection{Effectiveness and Efficiency}
To validate the effectiveness and efficiency of our proposed cross-modal skip-connected network, we conduct in-depth analysis on different stride values and various cross-modal fusion methods.

\subsubsection{Analysis of Stride for Skip}

\begin{figure}[t]
     \centering
     \includegraphics[width=0.5\textwidth]{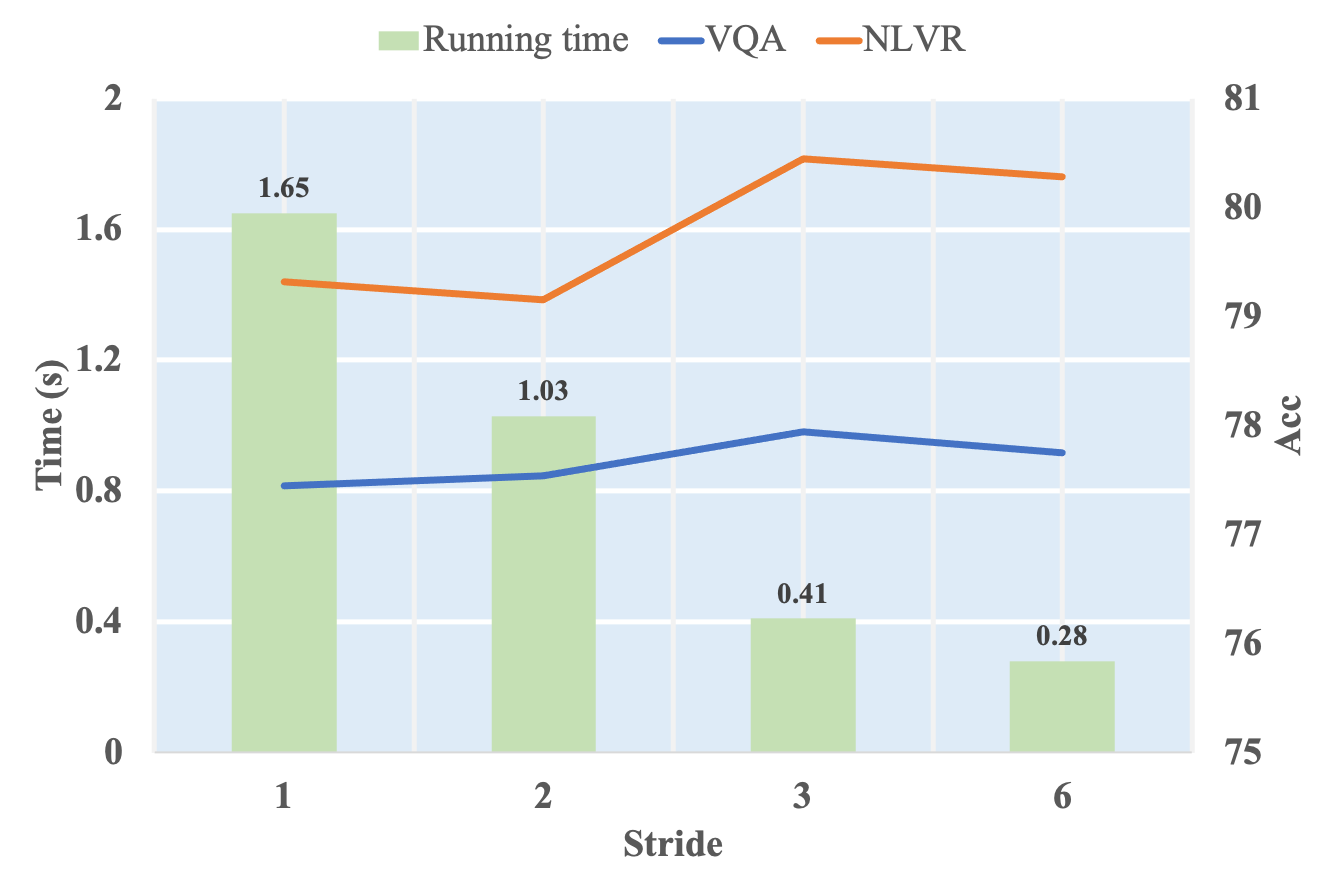}
     \caption{Results w.r.t different stride values  in cross-modal skip-connected network on running time and performance of VQA test-dev and NLVR2 test-P, where the running time is the total forward time of 100 samples.}
     \label{fig:analysis_stride}
\end{figure}

The stride \emph{S} is the key factor to control the effectiveness and efficiency tradeoff. Therefore, we further compare the running time and performance of different stride value \emph{S} in cross-modal skip-connected network on VQA and NLVR2 tasks. Specifically, we test four different stride values, which can be divisible by the total number of cross-modal fusion layers. The model is chosen as mPLUG{\tiny{ViT-B}} and all the other experiment settings are kept the same. As shown in Figure~\ref{fig:analysis_stride}, we can see that the larger \emph{S} is, the more efficient cross-modal fusion is, where the running time can be largely reduced from skipping the vision co-attention layers by 5$X$ times from $\emph{S}=1$ to $\emph{S}=6$. The performances of \modelname on both datasets gradually increases when $\emph{S}=3$, and slightly decreases later on. Compared with $\emph{S}=3$,  \modelname can achieve comparable performance at $\emph{S}=6$, while speeding up by nearly 30\%. Therefore, we set $\emph{S}=6$ on mPLUG{\tiny{ViT-L}} for faster pre-training.


\subsubsection{Analysis of Cross-modal Fusion}

\begin{figure}[t]
     \centering
     \includegraphics[width=0.5\textwidth]{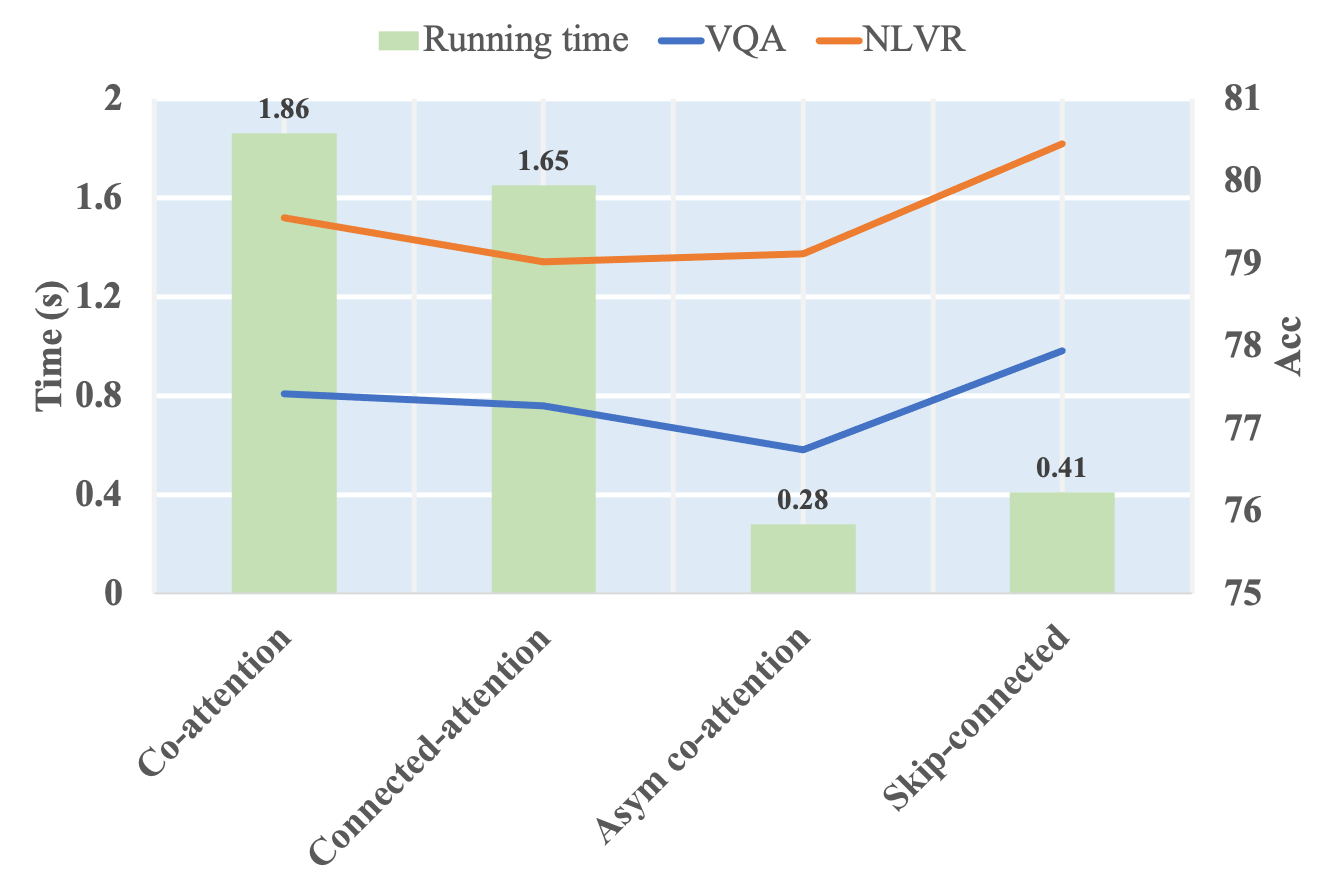}
     \caption{Results w.r.t different cross-modal fusions on running time and performance on VQA test-dev and NLVR2 test-P, where the running time is the total forward time of 100 samples.}
     \label{fig:analysis_fusion}
\end{figure}

We compare the effectiveness and efficiency of different cross-modal fusion variants in terms of running time and performance on VQA and NLVR2 tasks. Specifically, we pre-train \modelname with different cross-modal fusion network based on the same image encoder and text encoder. All the
pre-training settings and the number of fusion layers are kept the same as in the original \modelname pre-training. As shown in Figure~\ref{fig:analysis_fusion}, the fusion methods of co-attention and connected-attention both requires much more running time due to long visual sequence. Compared with the two fusion methods, our proposed skip-connected network is 4$X$ faster and obtain better performance on both datasets. We also compare it with the asymmetric co-attention used in BLIP ~\cite{li2021align,li2022blip} which only relies on the co-attention layers from images to text. Despite running slightly faster than the skip-connected network does, the asymmetric co-attention performs worse in accuracy on both datasets. The performance degradation is attributed to the information asymmetry and bias towards language, as shown in Section~\ref{section:vqa_analysis}.

%

\begin{table}[t]
\setlength\tabcolsep{0pt}
\centering
\begin{tabular}{lc}
\toprule
Model & Throughput (Samples/S) \\
\midrule
baseline & 124.0 \\
\ \ + BFloat16 & 182.7 \\
\ \ \ \ + Gradient Checkpoint & 238.2 \\
\ \ \ \ \ \ + ZeRO & \textbf{422.5} \\
\bottomrule
\end{tabular}
\caption{Training Throughput}
\label{table:large-scale}
\end{table}

\subsubsection{Large-scale Training}
Combining the techniques introduced in Section 4 has dramatically increased the training throughput. With the utilization of memory saving and accelerated training techniques, the throughput of \modelname improves 3$X$ more from 124 samples per second to 422 samples per second, as shown in Table \ref{table:large-scale}. 

\begin{table}[t]
\setlength\tabcolsep{3pt}
\centering
\begin{tabular}{lcccc}
\toprule
Model & In & Near & Out & Overall \\
\midrule
SimVLM$_{base}$\cite{wang2021simvlm}  & 83.2 & 84.1 & 82.5 & 83.5 \\
SimVLM$_{huge}$\cite{wang2021simvlm} & 101.2 & 100.4 & 102.3 & 101.4 \\
Oscar$\dagger$\cite{li2020oscar} & 85.4 & 84.0 & 80.3 & 83.4 \\
VinVL$\dagger$\cite{zhang2021vinvl} & 103.7 & 95.6 & 83.8 & 94.3 \\
SimVLM$_{huge}$$\dagger$\cite{wang2021simvlm} & 113.7 & 110.9 & 115.2 & 112.2  \\
\midrule
mPLUG & 86.34 & 81.5 & 90.49 & 84.02 \\
mPLUG$\dagger$  & \textbf{116.7}  & \textbf{113.75} & \textbf{117.0}  & \textbf{114.8} \\
\bottomrule
\end{tabular}
\caption{Image captioning results on NoCaps validation split (zero-shot and finetuned), and \{In, Near, Out\} refer to in-domain, near-domain and out-of-domain respectively. $\dagger$ denotes the models finetuned on COCO Caption dataset.}
\label{table:zero-shot-nocaps}
\end{table}

\begin{table}[t]
\centering
\begin{tabular}{l|cccc}
\toprule
\multicolumn{1}{l|}{\multirow{2}{*}{Model}}      &
     \multicolumn{2}{c}{TR} & \multicolumn{2}{c}{IR}  \\
    & R@1 & R@5 & R@1 & R@5 \\
\midrule
\multicolumn{5}{l}{\emph{Zero-Shot}} \\
\midrule
CLIP~\cite{radford2021learning}    &88.0& 98.7&68.7&90.6  \\
ALIGN ~\cite{jia2021scaling}     &88.6& 98.7&75.7&93.8  \\
FLIP ~\cite{yao2021filip} & 89.8 & 99.2 & 75.0 & 93.4 \\ 
Florence ~\cite{yuan2021florence} & 90.9 & 99.1 &76.7 &93.6\\
ALBEF$\dagger$   ~\cite{li2021align} & 94.1 & 99.5&82.8& 96.3 \\
BLIP$\dagger$ ~\cite{li2022blip} &94.8& 99.7  &84.9& 96.7                 \\
\midrule
mPLUG   & \textbf{93.0} &\textbf{99.5}& \textbf{82.2}&\textbf{95.8}  \\
mPLUG$\dagger$   & \textbf{95.8} &\textbf{99.8}& \textbf{86.4}&\textbf{97.6}  \\
\bottomrule
\end{tabular}
\caption{Zero-shot image-text retrieval results on Flickr30K. $\dagger$ denotes the models finetuned on COCO. }
\label{table:zero-shot-image-retrieval}
\end{table}

\begin{table}
\setlength\tabcolsep{1pt}
\small
\centering
\begin{tabular}{l|c|ccc}
\toprule
\multicolumn{1}{l|}{\multirow{2}{*}{Model}}  &
\multicolumn{1}{c|}{\# Pretrain} &
\multicolumn{3}{c}{MSRVTT-Retrieval} \\
  &  data & R@1 & R@5 & R@10  \\
\midrule
\multicolumn{5}{l}{\emph{Zero-Shot}} \\
\midrule
MIL-NCE~\cite{miech2020end} & How100M &9.9 & 24.0& 32.4\\
VideoCLIP~\cite{xu2021videoclip} & How100M &10.4 & 22.2& 30.0\\
VATT ~\cite{akbari2021vatt}& How100M, AudSet &- & -& 29.7\\
ALPRO ~\cite{li2021align_prompt}& W2M, C3M &24.1 &44.7 &55.4\\
VIOLET ~\cite{fu2021violet}& Y180M, W2M, C3M  &25.9 & 49.5 & 59.7\\
CLIP~\cite{radford2021learning} &WIT400M&26.0& 49.4& 60.7 \\
Florence ~\cite{yuan2021florence}& FLD900M &37.6 & 63.8 &72.6\\
BLIP $\dagger$ ~\cite{li2022blip} &129M  &43.3 & 65.6& 74.7\\
mPLUG &14M   & 38.1 & 59.2& 68.2\\
mPLUG $\dagger$ &14M   &\textbf{44.3}& \textbf{66.4} & \textbf{75.4}\\
\midrule
\multicolumn{5}{l}{\emph{Fine-Tuning}} \\
\midrule
VideoCLIP~\cite{xu2021videoclip} & How100M &30.9 & 55.4 & 66.8\\
ALPRO ~\cite{li2021align_prompt} & C3M, W2M & 33.9 & 60.7 & 73.2\\
VIOLET ~\cite{fu2021violet}& Y180M, C3M, W2M &34.5 & 63.0 & 73.4\\
\bottomrule
\end{tabular} 
\caption{Zero-shot video-language results on text-to-video retrieval on the 1k test split of the MSRVTT dataset. $\dagger$ denotes the models finetuned on COCO. Video datasets include HowTo100M~\cite{miech2019howto100m}, WebVid-2M(W2M)~\cite{bain2021frozen}, YT-Temporal-180M(   Y180M)~\cite{zellers2021merlot}. Image datasets include CC3M(C3M)~\cite{sharma2018conceptual}, FLD900M~\cite{yuan2021florence}, WIT400M~\cite{radford2021learning}. Audio datasets include AudioSet(AudSet)~\cite{gemmeke2017audio}.}
\label{table:video-retrieval-zeroshot}
\end{table}

\subsection{Zero-shot Transferability}
In this section, we examine the generalization of \modelname and compare the zero-shot result on two Vision-Language and three Video-Language tasks.
\subsubsection{Zero-shot Vision-Language Tasks}
The pretraining of \modelname adopts image-text contrastive and prefix language modeling tasks on large-scale image-text pairs. Thus, \modelname has zero-shot generalization ability in image-text retrieval and image captioning. \textbf{Image Caption}: First, we take the pretrained \modelname model and directly decode on NoCaps validation set without further finetuning. Following\cite{wang2021simvlm,li2022blip}, we 
feed a prefix prompt \emph{``A picture of''} into the text encoder to improve the quality of decoded captions. As shown in Table \ref{table:zero-shot-nocaps}, the zero-shot performance of \modelname is competitive with fully supervised
baselines such like Oscar and VinVL. With further finetuning on MSCOCO dataset, \modelname outperforms the SimVLM$_{huge}$, which use more pre-training image-text pairs and has larger model parameters. \textbf{Image-text Retrieval}: We perform zero-shot retrieval on Flickr30K. The result is shown in Table \ref{table:zero-shot-image-retrieval}, where zero-shot \modelname outperforms models (CLIP, ALIGN, Florence) pretrained with more image-text pairs. Following ~\cite{li2022blip}, we also evaluate zero-shot retrieval by the model finetuned on MSCOCO dataset. Table \ref{table:zero-shot-image-retrieval} shows that \modelname achieves better performance than the previous SOTA models.

\subsubsection{Zero-shot Transfer to Video-Language Tasks}
To evaluate the generalization ability of \modelname to Video-Language Tasks, we conduct zero-shot experiments on Video-text Retrieval, Video Caption and Video Question Answering. Following ~\cite{li2022blip}, we uniformly sample $n$ frames for each video ($n = 8$ for Retrieval, $n = 16$ for  QA, $n=8$ for Caption), and concatenate the frame features into a single sequence. \textbf{Video-text Retrieval}: We evaluate the \modelname models pretrained and further finetuned on the COCO-retrieval image-text dataset without any video pre-training or supervision. Table \ref{table:video-retrieval-zeroshot} shows that zero-shot \modelname can outperform the SOTA models pretrained on far more pretraining data (e.g., Florence, BLIP), and can even outperform models finetuned on the supervised video dataset without using temporal information (e.g., VideoCLIP, VIOLET); \textbf{Video Question Answering}: Following BLIP ~\cite{li2022blip}, We treat Video QA as an answer generation task and perform evaluation based on models finetuned on VQA. As shown in Table \ref{table:video-qa-zeroshot}, the zero-shot \modelname outperforms BLIP pretrained with more image-text pairs; \textbf{Video Caption}: We use a prefix prompt \emph{“A video of”} to improve the quality of decoded captions. Table \ref{table:video-qa-zeroshot} shows that zero-shot \modelname also achieves better performance than BLIP.

\begin{table}
\setlength\tabcolsep{4pt}
\small
\centering
\begin{tabular}{l|cc|c}
\toprule
\multicolumn{1}{l|}{\multirow{2}{*}{Model}}  &
MSRVTT-QA & MSVD-QA & VATEX-Cap \\
& Acc & Acc & CIDEr  \\
\midrule
\multicolumn{4}{l}{\emph{Zero-Shot}} \\
\midrule
VQA-T ~\cite{yang2021justask} &2.9&7.5&-\\
BLIP ~\cite{li2022blip} &19.2&35.2&37.4\\
mPLUG    &\textbf{21.1} &\textbf{37.2}& \textbf{42.0}\\
\bottomrule
\end{tabular} 
\caption{Zero-shot video-language results on Question-Answer and Caption tasks.}
\label{table:video-qa-zeroshot}
\end{table}


\section{Conclusion} 
This paper presents mPLUG, an effective and efficient VLP framework for both cross-modal understanding and generation. \modelname introduces a new asymmetric vision-language architecture with novel cross-modal skip-connections, to address two fundamental problems of information asymmetry and computation efficiency in cross-modal alignment. Pretrained on large-scale image-text pairs, \modelname achieves state-of-the-art performance on a wide range of vision-language tasks. \modelname also demonstrates strong zero-shot transfer ability when directly applied to multiple video-language tasks. Our work explores the cross-modal alignment with a newly-designed VLP architecture and we hope it can help promote future research on image-text foundation models. 




\bibliographystyle{acl_natbib}
\bibliography{references}


\input{appendix}

\end{document}

%% file: appendix.tex
\section{More Experiments Details}
\label{sec:app_expr}
\subsection{Downstream Task Details}
We evaluate mPLUG on the six downstream vision-language tasks. The hyperparameters that we use for finetuning on the downstream  tasks are listed in Table \ref{table:finetune-hyper}. Following ~\citep{li2021align}, all tasks adopt RandAugment, AdamW optimizer with a weight decay of 0.05 and a cosine learning rate schedule. We use an image resolution of 336 $\times$ 336, except for VQA where we use 504 $\times$ 504 images. For VQA and image captioning tasks, we also do an additional continue pre-training on 4M image-text pairs, which can bring about 0.2+ accuracy improvement. Next we introduce the dataset settings in detail.


\paragraph{VQA.} We conduct experiment on the VQA2.0 dataset ~\citep{goyal2017making}, which contains 83k/41k/81k images for training/validation/test. Following ~\citep{li2021align}, we use both training and validation splits for training, and incorporate additional training data from Visual Genome~\citep{krishna2017visual}.

\begin{table}
\setlength\tabcolsep{2pt}
\centering
\small
\begin{tabular}{l|ccc}
\toprule
Task  &  LR (ViT-L/BERT$_{base}$) & batch size & epochs  \\
\midrule
VQA & 2e-5/5e-6 & 1024 &  8 \\
Captioning$\dagger$ & 1e-5\&8e-7 & 256& 5 \\
Retrieval & 1e-5/2e-6 & 256& 5 \\
Visual Grounding & 2e-5/2e-6 & 512& 120 \\
NLVR2 & 5e-5/5e-6 & 256 & 15 \\
SNLI-VE & 2e-5 & 64&  5 \\
\bottomrule
\end{tabular} 
\caption{Finetuning hyperparameters for downstream tasks. $\dagger$ denotes two stages fine-tuning.}
\label{table:finetune-hyper}
\end{table}

\paragraph{Image Captioning.} We finetune on COCO’s Karpathy train split, and evaluate on COCO’s Karpathy test split and No-Caps validation split. Following~\cite{li2020oscar,wang2022OFA}, we first fine-tune \modelname with cross-entropy loss for 5 epochs with a learning rate of 1e-5 and a batch size of 256. Based on the fine-tuned model, we the fine-tune it with CIDEr optimization~\cite{scst} for extra 5 epochs with a smaller learning rate of 8e-7. During inference, we use beam search with a beam size of 10, and set the maximum generation length as 20.

\paragraph{Image-Text Retrieval.} We adopt the widely-used Karpathy split ~\citep{karpathy2015deep} for both COCO and Flickr30K. COCO contains 113/5k/5k images for train/validation/test, and Flickr30K contains 29k/1k/1k images for train/validation/test.

\paragraph{Visual Grounding.} We evaluate our method on three referring expression grounding datasets: RefCOCO, RefCOCO+~\cite{yu2016modeling} and RefCOCOg~\cite{mao2016generation}. The RefCOCO and RefCOCO+ datasets share 19K images and contain 142/141K queries. The RefCOCOg dataset contains 25K images and 95K queries.
To fully use training data, we first train the model with a mixed dataset with a learning rate of 2e-5. Then we continue fine-tuning the model on each dataset with a learning rate of 2e-6.

\paragraph{NLVR2 \& SNLI-VE.} We conduct experiment both on the official split~\cite{suhr2018corpus,snlive}.

\subsection{Pre-training Dataset Details}
Table \ref{table:pretraindata} shows the statistics of the 14M pre-training images with texts. 
\begin{table}[htbp]
\setlength\tabcolsep{4pt}
\centering
\small
\begin{tabular}{l|ccccc}
\toprule
  &  COCO & VG & SBU & CC3M & CC12M \\
\midrule
image & 113K & 100K & 860K & 3M & 10M  \\
text & 567K & 769K & 860K & 3M & 10M \\
\bottomrule
\end{tabular} 
\caption{Statistics of the pre-training datasets.}
\label{table:pretraindata}
\end{table}